\documentclass[10pt,twocolumn,letterpaper]{article}

\usepackage{iccv}              

\usepackage{cuted}

\definecolor{iccvblue}{rgb}{0.21,0.49,0.74}
\usepackage[pagebackref,breaklinks,colorlinks,allcolors=iccvblue]{hyperref}
\usepackage{multirow}
\usepackage{adjustbox}
\def\paperID{46} 
\def\confName{ICCV}
\def\confYear{2025}

\begin{document}
\title{CARDIUM: Congenital Anomaly Recognition with Diagnostic Images and Unified Medical records}



\author{
Daniela Vega\textsuperscript{1}
\qquad 
Hannah V. Ceballos\textsuperscript{1}
\qquad
Javier S. Vera\textsuperscript{1}
\qquad
Santiago Rodriguez\textsuperscript{1}\\
\qquad
Alejandra Perez\textsuperscript{1}
\qquad
Angela Castillo\textsuperscript{1}
\qquad
Maria Escobar\textsuperscript{1}\\
\qquad
Dario Londo\~no\textsuperscript{1,2}
\qquad
Luis A. Sarmiento\textsuperscript{1,2}
\qquad
Camila I. Castro\textsuperscript{1}\\
\qquad
Nadiezhda Rodriguez\textsuperscript{1,2}
\qquad
Juan C. Brice\~no\textsuperscript{1}
\qquad
Pablo Arbelaez\textsuperscript{1}
\vspace{0.1em}
\\
$^1$ Universidad de los Andes, Colombia
$^2$ Fundación Santa Fe de Bogotá, Colombia
\vspace{0.1em}
\\
\parbox{\textwidth}{\centering\tt\small\{d.vegaa,h.ceballos,j.verar,s.rodriguezr2,a.perezr20,a.castillo13,mc.escobar11,\\d.londono25,ansarmie,cami-cas,narodrig,jbriceno,pa.arbelaez\}@uniandes.edu.co}
}

\maketitle
\begin{abstract}
Prenatal diagnosis of Congenital Heart Diseases (CHDs) holds great potential for Artificial Intelligence (AI)-driven solutions. However, collecting high-quality diagnostic data remains difficult due to the rarity of these conditions, resulting in imbalanced and low-quality datasets that hinder model performance. Moreover, no public efforts have been made to integrate multiple sources of information, such as imaging and clinical data, further limiting the ability of AI models to support and enhance clinical decision-making. To overcome these challenges, we introduce the \textbf{C}ongenital \textbf{A}nomaly \textbf{R}ecognition with \textbf{D}iagnostic \textbf{I}mages and \textbf{U}nified \textbf{M}edical records (CARDIUM) dataset, the first publicly available multimodal dataset consolidating fetal ultrasound and echocardiographic images along with maternal clinical records for prenatal CHD detection. Furthermore, we propose a robust multimodal transformer architecture that incorporates a cross-attention mechanism to fuse feature representations from image and tabular data, improving CHD detection by 11\% and 50\% over image and tabular single-modality approaches, respectively, and achieving an F1-score of 79.8 ± 4.8\% in the CARDIUM dataset. We will publicly release our dataset and code to encourage further research on this unexplored field. Our dataset and code are available at \url{https://github.com/BCV-Uniandes/Cardium} , and at the project website \url{https://bcv-uniandes.github.io/CardiumPage/}.
\end{abstract}
   
\begin{figure}[htbp]
    \centering
    \includegraphics[width=0.38\textwidth]{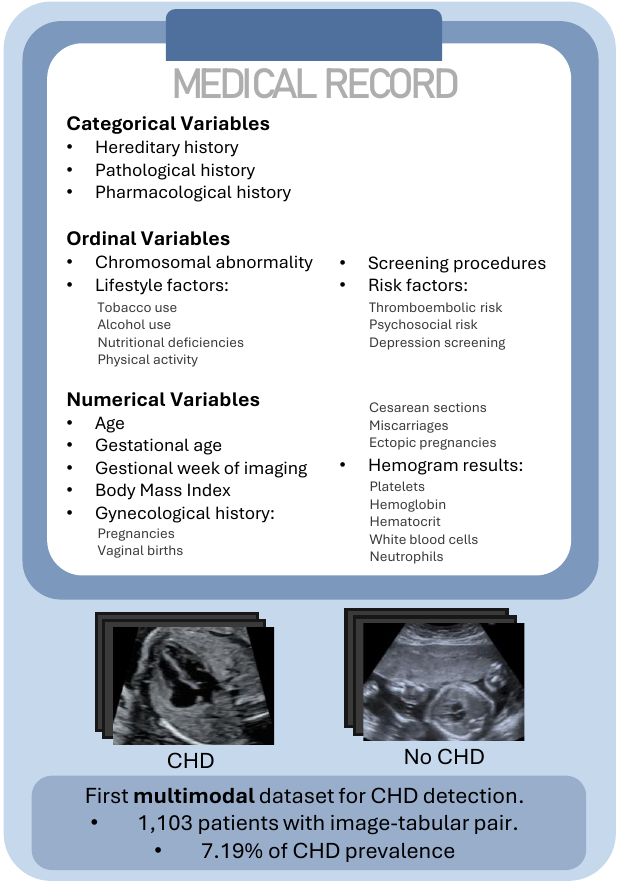}
    \caption{\textbf{Overview of the CARDIUM dataset}. The CARDIUM dataset includes diagnostic images from 1,103 patients and 26 physiological variables from the mother's clinical record.}
    \label{fig:overview_figure}
\end{figure}

\section{Introduction}
\label{sec:intro}

Congenital Heart Diseases (CHDs) are structural abnormalities of the heart and blood vessels that develop during fetal growth and are the leading cause of infant mortality \cite{ottaviani2016}. Prenatal detection through ultrasound and echocardiographic imaging is crucial to improving clinical outcomes. Yet, detection rates can be as low as 30\%, particularly in low- and middle-income countries, due to limited access to specialists and equipment \cite{arnaout2020}, \cite{elshazali2022}.

Artificial Intelligence (AI) offers the potential to reduce these disparities and improve prenatal diagnosis \cite{jone2022} by assisting specialists in recognizing cardiac abnormalities. However, the unique characteristics of these conditions, along with the sensitivity involved in working with fetal data, introduce significant challenges.

First, CHDs are extremely rare, affecting approximately 8 in every 1,000 live births globally each year, which makes it difficult to collect extensive and diverse datasets \cite{van2011}. Moreover, the small size of the fetal heart and the fetus's constant movement make it challenging to acquire clear diagnostic images \cite{liu2024}. As a result, existing datasets are often imbalanced and of low quality, which limits the ability of AI models to learn robust and generalizable patterns. Integrating clinical data could help compensate for the scarcity and imbalance of imaging datasets; however, such approaches remain largely unexplored.

Second, fetal data is highly sensitive, requiring strict regulations and extensive approvals for collection and sharing. Consequently, creating publicly available CHD datasets is very challenging. Nevertheless, access to public datasets is essential for meaningful progress in automated CHD detection, as it ensures the reproducibility of AI models, encourages collaboration, and accelerates the development of more effective diagnostic methods.

To address these limitations, we propose two key contributions in this paper. First, we introduce the \textbf{C}ongenital \textbf{A}nomaly \textbf{R}ecognition with \textbf{D}iagnostic \textbf{I}mages and \textbf{U}nified \textbf{M}edical records (CARDIUM) dataset, the first publicly available multimodal dataset for prenatal CHD detection. This dataset combines echocardiographic and ultrasound images with maternal clinical data, enabling a more comprehensive analysis of CHD risk, while facilitating open research and fair comparisons between methods. Second, we present the CARDIUM model, a multimodal transformer that achieves promising results on our dataset, establishing a baseline for future studies and encouraging advancements in prenatal CHD diagnosis. We will make our dataset and code publicly available to promote open research.

\section{Related Work}
\subsection{Deep Learning Algorithms for Congenital Heart Disease Detection}

The rapid advancements of deep learning have led to significant progress in AI-based methods for prenatal CHD detection. For instance, Arnout \etal~\cite{arnaout2020} trained a ResNet, achieving an AUC of up to 99\% across four datasets. Qiao \etal~\cite{qiao2022} used a residual CNN, reaching 93\% accuracy in four-chamber fetal images. Moreover, Nurmani \etal~\cite{nurmaini2022} employed a DenseNet21, achieving 92\% inter-patient and 100\% intra-patient accuracy.  Despite these promising results, all methods rely on private datasets and, except for \cite{arnaout2020}, lack publicly available code, hindering reproducibility and fair comparison. Furthermore, none incorporate multimodal data, limiting their ability to replicate real-world clinical practice \cite{krones2025}. Our approach addresses these limitations by introducing the first public multimodal dataset for CHD detection, along with an open-source multimodal baseline model.

\subsection{Multimodal Models}
In clinical practice, physicians rely on multiple data types, including medical images and clinical records, to make accurate diagnoses. Some multimodal models combining imaging and tabular data have been explored for other diagnostic tasks. Hager \etal~\cite{hager2023} combine imaging and tabular data in a contrastive multimodal learning (MMCL) framework, achieving AUCs of 73.76\% for predicting coronary artery disease risk and 76.60\% for predicting myocardial infarction risk on the UK Biobank dataset \cite{ukbiobank2018}. More recently, Du \etal~\cite{du2024} introduced Tabular-Image Pre-training (TIP), which improves on MMCL by combining image–tabular contrastive learning, masked tabular reconstruction, and image–tabular matching, achieving AUCs of 86.43\% and 85.58\% on the same datasets. Despite these promising results, multimodal approaches for CHD detection remain largely unexplored. Moreover, both methods struggle with class imbalance, limiting their clinical applicability in scenarios like CHD diagnosis, where positive cases are far less common than negative ones. Although there are other studies on multimodal diagnostic models, these do not use tabular information and images as input modalities, and some require further adjustments to be comparable. In this context, the CARDIUM model emerges as the first multimodal approach for CHD detection, incorporating strategies to address class imbalance and enhance robustness for clinical use.


\subsection{Datasets}
Automated diagnosis of CHD remains constrained by limited and inaccessible datasets.  ImageCHD \cite{xu2021} is the first open-access dataset for CHD classification; however, it is restricted to postnatal cases, underscoring the need for prenatal recognition datasets. Moreover, ImageCHD relies exclusively on imaging data, whereas real-world diagnoses also incorporate clinical information. CARDIUM represents the first multimodal dataset specifically designed for prenatal CHD classification, promoting the development and evaluation of novel algorithms and enabling significant advances in this field.

\begin{figure*}[h]
    \centering
    \includegraphics[width=0.87\textwidth]{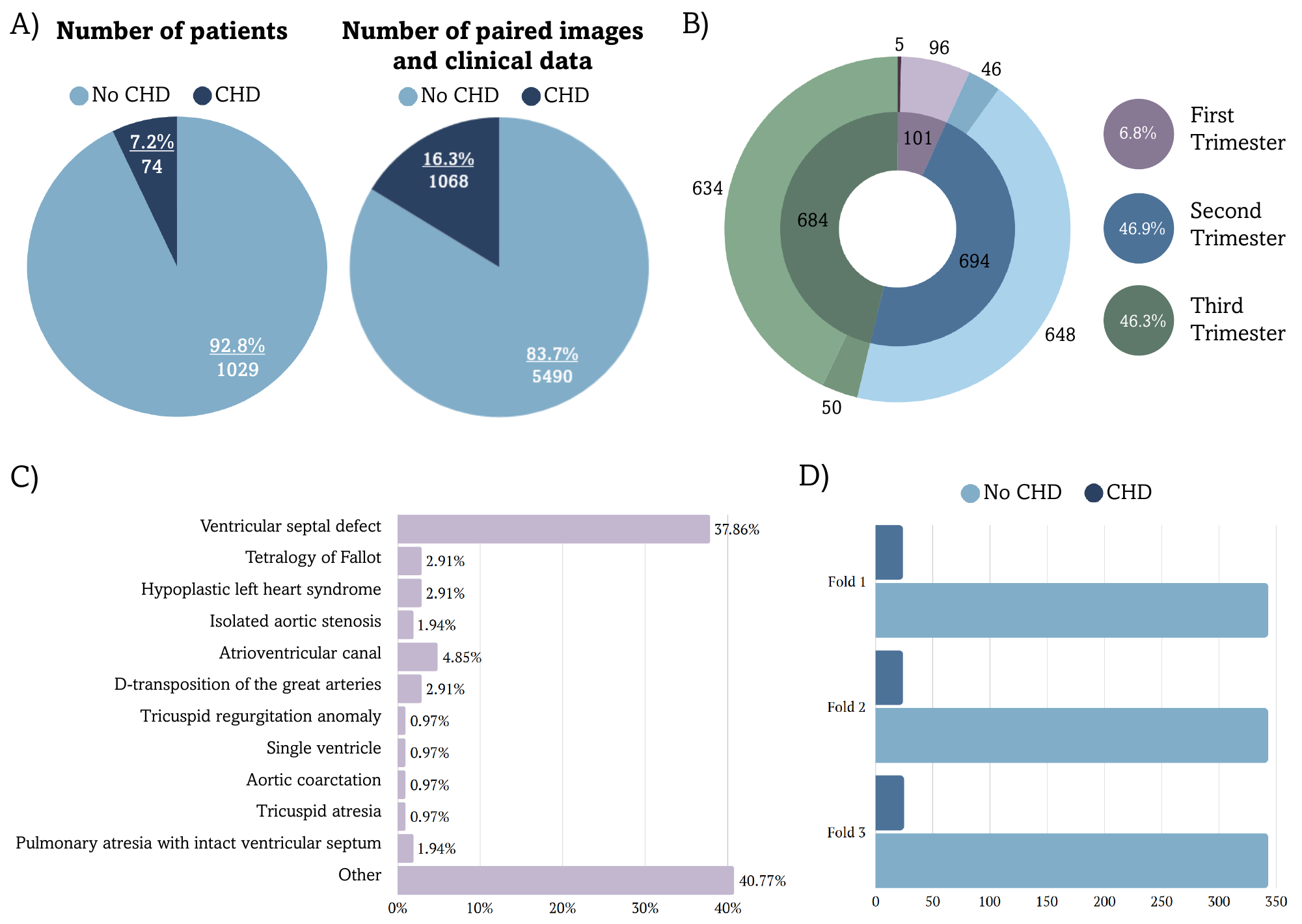}
    \caption{\textbf{CARDIUM dataset statistics}. (A) Number of patients with and without CHD (left) and number of images corresponding to patients with and without CHD (right). (B) Trimester distribution: The inner circle represents the overall number of patients with images from the first, second, and third trimesters, while the outer circle distinguishes between positive (darker) and negative (lighter) cases. On the right, we present the percentage of the total dataset corresponding to each gestational period. (C) Distribution of different types of CHDs present in our dataset. (D) Number of patients with and without CHD per fold.}
    \label{fig:statistics}
\end{figure*}

\section{CARDIUM Dataset}
\label{sec:cardium}

We present the CARDIUM dataset, the first multimodal dataset for prenatal CHD detection. CARDIUM combines the mother's clinical record with echocardiographic and ultrasound images, providing complementary diagnostic modalities that collectively contribute to a holistic understanding of the fetus's physiological state. The dataset was constructed through a retrospective study on Colombian women, with data collected between 2013 and 2024.

\subsection{Image Collection}

We acquired 2D echocardiographic and ultrasound images using Voluson E6/E8/E10 systems (GE Healthcare, Austria), following established protocols \cite{Moon2023}. For each examination, a CHD specialist captured the standard four cardiac views included in routine fetal ultrasound evaluations: the four-chamber view, the three-vessel trachea view, the left ventricular outflow view, and the right ventricular outflow view. These views provide different perspectives of the fetus's heart, offering crucial anatomical insights for CHD detection. 

After image acquisition, an expert echographer reviewed all images and discarded those that were considered inconsistent or of very low quality. We retained all images approved by the echographer, including multiple images of the same view, although not all examinations contained all four views. As a result, each patient had more than one image. 

Color and power Doppler with high-definition flow enhanced image quality and vascular detail. 

\subsection{Medical Records Collection}

We extracted categorical and numerical variables from the mother's medical records by converting event-based notes into a tabular format. We selected categories that capture essential maternal and fetal health indicators, providing relevant physiological context. We also confirmed that the chosen categories were available across most clinical records. Figure \ref{fig:overview_figure} showcases all the variables included in the dataset, along with their corresponding data type (categorical, ordinal, and numerical).

Since the available clinical data and ultrasound images were not collected on the same day, we consolidated all available medical records from the duration of each pregnancy to capture a broader clinical context. Specifically, we aggregate all clinical events for a given patient into a single tabular entry, which often contains multiple values for most numerical variables. For variables that remain stable throughout pregnancy, such as gynecological history, we retained only one value, as these do not change across events. We also retained a single value per patient for ordinal variables. In binary fields, which include chromosomal abnormalities, screening procedures, and lifestyle factors, we assigned a value of 1 ("yes") if any record indicated a positive case. For ordinal scale fields, such as risk factors, we selected the highest reported level across all records (low, intermediate, or high). This approach ensured that clinically relevant risks were not underestimated due to variability in timing or documentation. For categorical variables, we included all categories recorded across the available medical records. Pathological, hereditary, and pharmacological histories contained 74, 43, and 50 unique categories, respectively.

Although some clinical variables were recorded after the ultrasound images were acquired, all data were collected during the same pregnancy, ensuring they reflected a consistent clinical context. Most variables, such as pathological history, hereditary history, and risk factors, remain stable throughout pregnancy or are more reliably documented during later visits. Including these data provides a comprehensive and accurate clinical profile that closely reflects the type of information typically available alongside ultrasound and echocardiographic imaging.

\begin{figure*}[h!]
    \centering
    \includegraphics[width=0.95\textwidth]{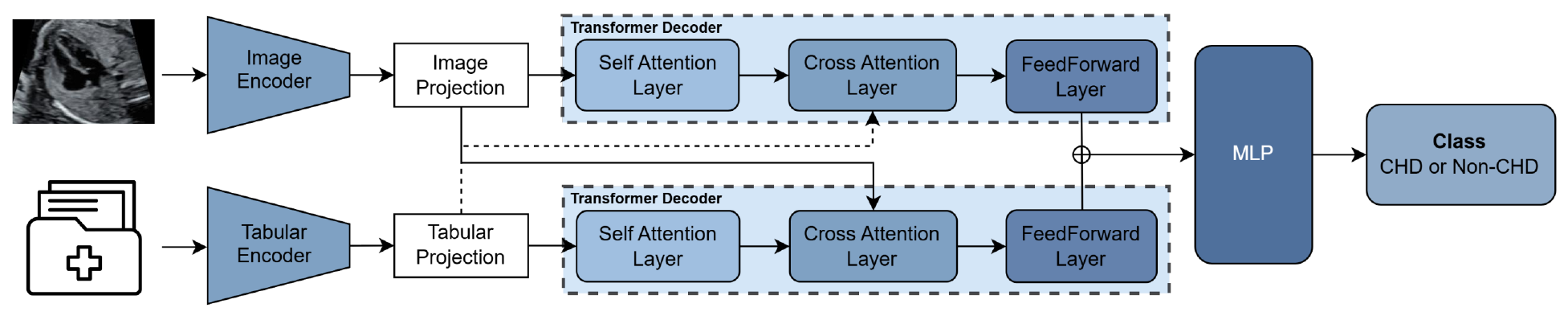}
    \caption{\textbf{Overview of the CARDIUM model}. We process image and tabular data through modality-specific encoders, \( E_I \) and \( E_T \), to obtain distinct embeddings. We pass these embeddings through transformer decoder layers, where modality fusion occurs in the cross-attention layer. Then, we concatenate the fused representations and process them through a Multi-Layer Perceptron (MLP) to classify cases as CHD or non-CHD.}
    \label{fig:overview_figure_model}
\end{figure*}

\subsection{Dataset Statistics}
Figure \ref{fig:overview_figure} provides an overview of our dataset, highlighting selected variables and imaging examples. The study involved a population of 1,103 patients with either obstetric echocardiographic or ultrasound images and associated clinical records available at the Fetal-Maternal Medicine Unit. In cases where multiple images were available for a single patient (\eg, from different visits), we linked all images to a single tabular record that consolidated information from all clinical events. All patients were required to be over 18 years old, and those with twin pregnancies were excluded from the study. The cohort had a mean age of 34.86 ± 4.92 years. Given the relatively low prevalence of CHD, gathering a sufficient number of positive cases required a significant effort. However, we achieved a CHD prevalence of 7.19\%, which is higher than the approximately 1\% observed in the general population \cite{van2011}. Furthermore, since each patient could have more than one image, the total number of images is 6558, with 16.3\% corresponding to positive patients and 83.7\% corresponding to negative patients. These statistics are depicted in Figure \ref{fig:statistics}A.

We collected data from various stages of pregnancy, as shown in Figure \ref{fig:statistics}B. This figure illustrates the number of patients, both CHD-positive and CHD-negative, in each trimester, along with the percentage of the total dataset corresponding to each gestational period. Additionally, Figure \ref{fig:statistics}C presents the distribution of CHD types in the CARDIUM dataset. The dataset contains images from 11 of the most frequent CHD types worldwide, with a 12th category labeled "Other" for less frequent conditions \cite{AHA2022}.

We divided the dataset into three cross-validation folds to ensure robust evaluation and better assess the model's generalization. Stratified sampling preserved the CHD and non-CHD proportions across folds, as shown in Figure \ref{fig:statistics}D, ensuring each fold accurately reflects the overall distribution in the CARDIUM dataset.

\subsection{Data Privacy and Ethical Approval}

To ensure patient privacy, we implemented strict anonymization protocols by assigning unique anonymized IDs and removing all sensitive information from the tabular data. Images were securely stored on the REDCap platform, which provides robust data protection and adheres to ethical and legal standards. The research protocol received approval from the Institutional Review Board (IRB) in accordance with international ethical guidelines.


\subsection{Tabular Data Preprocessing}
We establish a dataset preprocessing pipeline with two key components: numerical and categorical data refinement and categorical variable encoding.

\subsubsection{Numerical Data Refinement}
For numerical data refinement, we standardize the units of all numerical variables to ensure consistency across medical records. After unit standardization, we rectify any out-of-bounds values and apply z-score normalization to all numerical features (mean of 0 and standard deviation of 1).

\subsubsection{Categorical Data Refinement} We first correct typographical errors using a combination of automated scripts and manual review. We also review categorical values and standardize the names of diseases and medications, as naming conventions often vary between clinical records despite referring to the same underlying category (\eg, \textit{progesterone}, \textit{progesterone intravaginal}, \textit{Progendo}). 

We then group categorical variables based on semantic similarity to reduce the number of unique entries in the pathological, hereditary, and pharmacological history fields. For example, terms such as \textit{vaginitis}, \textit{candidiasis}, and \textit{acute vaginitis} were all grouped under the broader category of \textit{vaginal infections}. Finally, we combine categories with fewer than four occurrences into an "Others" label. 

\subsection{Evaluation Metrics} Given the imbalanced nature of the CARDIUM dataset, we propose evaluating the model's performance using a three-fold cross-validation strategy. During training, we treat each image as an individual sample to maximize the available data. However, for inference, we compute metrics on a per-patient basis by averaging the outputs from all corresponding images, aligning the evaluation process more closely with clinical practice. To assess the model’s ability to identify CHD cases, we report the F1-score, precision, and recall for the CHD class, along with the Area Under the Receiver Operating Characteristic Curve (AUC) to measure overall performance.
\section{CARDIUM Multimodal Model Architecture}

The CARDIUM model is a novel multimodal framework that leverages a dual cross-attention mechanism to capture intricate dependencies between imaging and clinical data. Figure \ref{fig:overview_figure_model} illustrates the CARDIUM model architecture. 

Given an image \( I \in \mathbb{R}^{C \times H \times W} \) and encoded tabular data \( T \in \mathbb{R}^{n} \), where \( n \) represents the number of tabular features, we process each type of data through its corresponding encoder. The image \( E_I \) and tabular \( E_T \) encoder transform their inputs into
modality-specific embeddings \( z_I \in \mathbb{R}^{1 \times D_I} \) and \( z_T \in \mathbb{R}^{1 \times D_T} \), respectively. Here,
\( D_I \) and \( D_T \)  denote the embedding dimensions for each modality. We map both representations into a shared representation space with dimension \( D \) and input them into the multimodal fusion architecture for inter-modality learning. Full implementation details can be found in Appendix \ref{sec:implemetation_details}.

\begin{table*}[t]
\centering
\caption{CHD detection results on the CARDIUM dataset for modality-specific variants of our model.}
\begin{adjustbox}{width=0.74\textwidth}
\setlength{\tabcolsep}{3pt}
\small
\begin{tabular}{@{}c||c||cccc@{}}
\toprule
\textbf{Images} & \textbf{Clinical Data} & \multicolumn{1}{c} \scriptsize {\textbf{CHD F1 Score}} & \multicolumn{1}{c}{\textbf{CHD Precision}} & \multicolumn{1}{c}{\textbf{CHD Recall}} & \multicolumn{1}{c}{\textbf{AUC}} \\ 
\midrule
\small\checkmark & \small \checkmark & \scriptsize \textbf{0.798 ± 0.048} & \scriptsize \textbf{0.876 ± 0.173} & \scriptsize \textbf{0.757 ± 0.104} & \scriptsize \textbf{0.974 ± 0.012} \\ 
\small \checkmark  &  & \scriptsize 0.689 ± 0.066 & \scriptsize 0.659 ± 0.135 & \scriptsize 0.742 ± 0.119 & \scriptsize 0.955 ± 0.0154 \\ 
& \small \checkmark  & \scriptsize 0.294 ± 0.019  & \scriptsize 0.192 ± 0.019 & \scriptsize 0.634 ± 0.049 & \scriptsize 0.794 ± 0.028 \\  
\bottomrule
\end{tabular}
\end{adjustbox}
\label{tab:results}
\end{table*}

\begin{figure*}[t]
\centering
\includegraphics[width=0.92\textwidth]{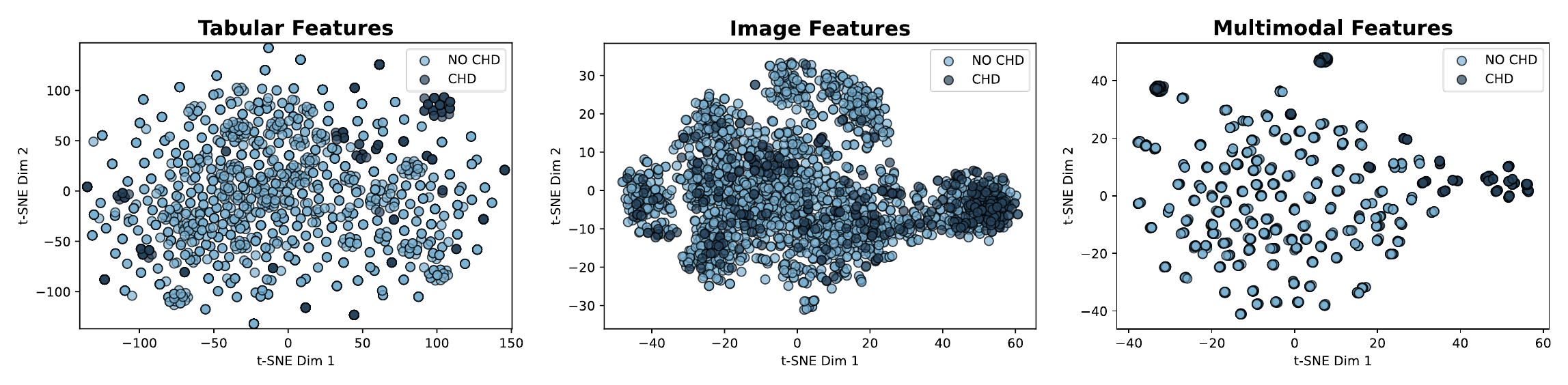}
\caption{Feature distributions for CHD and non-CHD cases across intermediate outputs of our multimodal model on the CARDIUM dataset. (A) Tabular encoder. (B) Image encoder. (C) Final multimodal module. In plot (C), the point density appears lower compared to plots (A) and (B); however, this lower density is due to overlapping points. Feature distributions are visualized using t-SNE \cite{tSNE}.}
\label{fig:results}
\end{figure*}

\subsection{Image Module} 

\subsubsection{Image Encoder}
We experiment with various architectures for the image encoder, including both Convolutional Neural Network (CNN)-based and transformer-based models. The best results are achieved by fine-tuning a Vision Transformer (ViT) \cite{dosovitskiy2020} model pre-trained on ImageNet. This ViT configuration consists of twelve layers and six attention heads, with dropout rates for the transformer path and classification head set at 0.3 and 0.2, respectively.

\subsection{Tabular Module} 

\subsubsection{Categorical Variables Encoding} To capture the relationship between categorical variables and the target outcome, we use Weight of Evidence (WoE) encoding, a Bayesian encoding technique that is inherently target-aware \cite{WoE}. This method assigns each category a numerical value based on the log-odds ratio of positive to negative class observations, effectively quantifying how informative a category is in predicting the target. To prevent data leakage, we employ a five-fold cross-encoding strategy, in which the encoding for each fold is computed using data from the remaining four folds. This ensures that the encoding of a category is not influenced by the target labels in the fold being evaluated. For further details on the encoding strategy, see Appendix \ref{sec:woeencoding}.

\subsubsection{Tabular Encoder}
To effectively encode tabular features, we modify the transformer architecture to process both numerical and encoded categorical data. In this approach, we treat each feature as an individual token. First, we project these tokens into a higher-dimensional space and process them through a two-layer transformer encoder with eight attention heads. Next, we flatten the output and map it to an embedding space. This design allows the model to capture complex dependencies between features, resulting in a rich tabular representation suitable for multimodal fusion.


\subsection{Multimodal Interaction Module} For multimodal representation learning, we employ a transformer decoder architecture to capture both intra- and inter-modality relationships through self-attention and cross-attention mechanisms. Let \( z_I \in \mathbb{R}^{1 \times d} \) and \( z_T \in \mathbb{R}^{1 \times d} \) denote the feature representations extracted from the image and tabular encoders, respectively. For each modality, we treat each batch of patients as a sequence, where each patient’s feature vector is treated as a token in this sequence. This design enables the model to learn contextual dependencies among patients within the batch, allowing it to assign adaptive importance to different features and effectively capture relationships across both modalities.

The interaction module consists of two parallel stacks of \( L \) transformer decoder layers, each comprising self-attention,  cross-attention, and feedforward layers. Each single-modality representation first undergoes self-attention, allowing the model to capture intra-modality dependencies. The self-attention mechanism enables the model to analyze complex relationships among features within a single modality and to capture dependencies between patients within the same modality. 

Subsequently, the output of the self-attention layer interacts with the representation from the opposite modality through the cross-attention mechanism. Here, the key and value matrices are derived from the encoder’s output of the opposite modality, while the query originates from the self-attention output. The cross-attention mechanism allows for effective information exchange between modalities, helping each representation refine itself by using complementary features from the other modality. By dynamically re-weighting features, cross-attention highlights critical diagnostic patterns that may not be as apparent in a single modality. 

We process the output from the cross-attention through a feedforward layer. Finally, we concatenate the refined features from both modalities and pass them through a three-layer Multilayer Perceptron (MLP) for classification.

\section{Results and Discussion}
\subsection{Multimodal CHD Detection Results}

Table \ref{tab:results} displays the overall performance of our model on the CARDIUM dataset, highlighting the effects of using one or both modalities. The results demonstrate that combining fetal echocardiography and ultrasound images with clinical data enhances performance by 11\% compared to using images alone and by 50\% compared to using clinical data alone. These findings align with real-world clinical practice, where, although fetal echocardiography is the primary diagnostic tool, physicians benefit significantly from maternal-specific clinical information to improve diagnostic accuracy.

Figure \ref{fig:results} further illustrates the impact of multimodal integration on feature representation. Plots (A) and (B) display the feature distributions from the tabular and image encoders, respectively, while plot (C) presents the fused multimodal representation. After fusion, class clusters become more compact and distinct, enhancing the model’s ability to differentiate between CHD and non-CHD cases. This visualization highlights the transformer decoder's effectiveness in capturing both intra- and inter-modality relationships, as well as the complementarity between data modalities. Consequently, multimodal fusion creates a more discriminative and structured feature space, enhancing CHD detection accuracy.

\subsection{Trimestral Model Performance}

We evaluate our model's performance separately on data from the first, second, and third trimesters. This allows us to assess the model's ability to detect CHD at different stages of pregnancy. As shown in Table~\ref{tab:comparison_trimesters}, the model performs best with data from the third trimester and shows the most difficulties with data from the first trimester. However, it is important to note that only five CHD-positive cases are available in the first trimester, making it difficult to draw definitive conclusions about the model's effectiveness at this early stage.

\begin{table}[h!]
\centering
\caption{Comparison of our model's performance on data collected during the first, second, and third trimesters of pregnancy.}
\begin{adjustbox}{width=0.47\textwidth}
\label{tab:comparison_trimesters}
\begin{tabular}{l||ccc}
\toprule
\textbf{Trimester} & \textbf{CHD F1 Score} & \textbf{CHD Precision} & \textbf{CHD Recall} \\ \midrule
First & 0.222 ± 0.314 & 0.333 ± 0.471 & 0.167 ± 0.236 \\
Second & 0.603 ± 0.092 & 0.701 ± 0.212 & 0.556 ± 0.101 \\
Third  & 0.732 ± 0.072 & 0.825 ± 0.127 & 0.669 ± 0.074 \\ \bottomrule
\end{tabular}
\end{adjustbox}
\end{table}

These findings align with clinical expectations, as CHD detection improves in later gestational stages when cardiac anomalies become more visible \cite{ULSCR2025} \cite{Hashiramoto2025}. However, strong performance during early stages remains critical, given the significant impact of early diagnosis on the baby’s prognosis. The promising results from the second trimester underscore the potential of such tools for early CHD detection and highlight the need for additional early-stage data to improve the model’s ability to identify CHD during the initial phases of fetal development. Furthermore, the model achieves higher overall performance when evaluated on the full dataset (79.8\% ± 4.8\%), emphasizing the importance of comprehensive data for robust CHD detection.


\subsection{Performance on Image Only Data}

We evaluate the CARDIUM model’s ability to detect CHD using only images from patients without available clinical records in the hospital’s database. For this evaluation, we collected ultrasound and echocardiographic images from 11 patients with CHD and 113 patients without CHD, resulting in 144 CHD images and 767 non-CHD images. We performed inference on these images, achieving an F1-score of 0.8528 ± 0.106. This result demonstrates that the CARDIUM model can detect CHD effectively in unimodal contexts.



\subsection{Generalization Experiments}

To evaluate our multimodal model's generalization capability, we compare its performance with state-of-the-art methods using a publicly available ultrasound fetal dataset \cite{burgos2020}, which includes maternal-fetal screening images from six anatomical planes. We use the same training/test split as proposed in \cite{burgos2020} to ensure a fair and consistent comparison. We evaluate performance using ViT-Small, our multimodal approach trained from scratch, and our multimodal model pre-trained on CARDIUM. Implementation details and the modifications made to adapt our model for a unimodal multiclass classification task are provided in Appendix \ref{sec:fetal-dataset}.


\begin{table}[h!]
\centering
\caption{Generalization Results on the Fetal-Planes-DB dataset \cite{burgos2020}.}
\begin{adjustbox}{width=0.45\textwidth}
\begin{tabular}{@{}c||c@{}}
\toprule
\textbf{Model}                            & \textbf{F1 Score} \\ \midrule
ViT Small & 0.900    \\ \
CARDIUM model (ours) & 0.914    \\
CARDIUM model (ours) pretrained on CARDIUM dataset & 0.918    \\ 
\midrule
\textbf{MedMamba-B \cite{yue2024}}                      & \textbf{0.933}    \\
VMamba-B \cite{liu2024_2}                        & 0.927    \\
Swin Transformer-B \cite{liu2021}              & 0.854    \\
ConvNext-B \cite{liu2022}                      & 0.855    \\
EfficientNetV2-B \cite{tan2021}                & 0.885    \\

\bottomrule
\end{tabular}
\end{adjustbox}
\label{tab:SOTA}
\end{table}

The results for ViT-Small, our multimodal approach trained from scratch, and our multimodal model pre-trained on CARDIUM are summarized in the top section of Table \ref{tab:SOTA}. These results show a gradual improvement in F1-score, with the multimodal approach outperforming ViT-Small and further improvements resulting from pre-training on CARDIUM. This behavior suggests that our multimodal framework enhances image representations, improving classification even in unimodal settings. Additionally, pre-training on CARDIUM consistently increased performance on an external ultrasound dataset, highlighting the dataset’s rich and transferable features. 

Moreover, the results show that although MedMamba-B achieved the best results, our approach outperforms leading methods such as Swin Transformer, EfficientNet V2, and ConvNext, indicating effective generalization across distinct ultrasound datasets and tasks.


\begin{table}[h!]
\centering
\caption{Comparison of our multimodal model with multimodal state-of-the-art approaches on the CARDIUM dataset.}
\begin{adjustbox}{width=0.33\textwidth}
\label{tab:comparison_sota}
\begin{tabular}{c||c}
\toprule
\textbf{Model} & \textbf{F1 Score} \\ \midrule
\textbf{CARDIUM model (ours)} & \textbf{0.798 ± 0.048}\\
TIP \cite{du2024}                    & 0.459 ± 0.027 \\
MMCL \cite{hager2023}                    & 0.349 ± 0.090 \\ \bottomrule
\end{tabular}
\end{adjustbox}
\end{table}
\renewcommand{\arraystretch}{1.0} 

\subsection{Comparison with SOTAs}

We compare our model’s performance with two state-of-the-art multimodal methods for binary classification using tabular and imaging data: MMCL \cite{hager2023} and TIP \cite{du2024}. Both methods were evaluated on our dataset using the same data split used in CARDIUM model. The implementation details applied to each model are described in Appendix \ref{sec:sota}.

Table \ref{tab:comparison_sota} presents the results of TIP and MMCL evaluated on CARDIUM. Both models underperformed compared to our model, which may be attributed to the significant class imbalance present in our dataset.  As noted in MMCL \cite{hager2023}, contrastive learning struggles in scenarios involving imbalanced binary classifications, and both TIP and MMCL rely on contrastive learning strategies. These results emphasize that our multimodal approach, along with the strategies we employ to handle class imbalance, is highly effective, providing a distinct advantage in real-world clinical situations where negative cases are much more common than positive ones.

\subsection{Ablation Experiments}

\subsubsection{Ablation on Training on Half the Data}

To assess the impact of data quantity on the performance of the CARDIUM model, we train the model using half of the CARDIUM dataset and evaluate it on the full test split. The results reveal a 13\% decrease in F1-score when only half of the data is used for training, highlighting the critical role data quantity plays in AI model performance. The CARDIUM model demonstrates higher performance when trained on a larger dataset, underscoring the importance of continually increasing dataset size to enhance CHD detection accuracy. See Appendix \ref{sec:ablation_half_data} for implementation details.



\subsubsection{Ablation on Different Multimodal Modules} 

We implement and evaluate several multimodal fusion strategies. \textit{MLP-Fusion} concatenates modality features and processes them with an MLP. \textit{Transformer Encoder Fusion} concatenates features and processes them with a transformer encoder. \textit{Transformer Decoder Fusion} processes image features with a transformer decoder and integrates tabular features via cross-attention. Finally, \textit{Transformer Encoder with Cross-Attention Fusion} encodes each modality separately and fuses them using cross-attention. See Appendix \ref{sec:multimodal_fusion} for further details.

Table \ref{tab:multimodal_models} presents the performance of these strategies compared to our final architecture. Our model outperforms all other approaches by at least 11\%, highlighting its effectiveness in capturing complex multimodal relationships. Self-attention enables the model to extract rich intra-modality dependencies, while the dual cross-attention strategy enhances feature representation through modality interaction, resulting in stronger fusion and improved performance.

\begin{table}
\centering
\caption{Results of different modality integration strategies.}
\begin{adjustbox}{width=0.45\textwidth}
\begin{tabular}{@{}c||l@{}}

\toprule
\textbf{Multimodal Module} & \multicolumn{1}{c}{\textbf{F1 Score}} \\ 
\midrule
MLP Fusion & 0.454 ± 0.067 \\
Transformer Encoder Fusion & 0.686 ± 0.086 \\
Transformer Decoder Fusion & 0.607 ± 0.091 \\
Transformer Encoder with Cross Attention Fusion & 0.681 ± 0.048 \\
\textbf{Double Transformer Decoder Fusion (ours)} & \textbf{0.798 ± 0.048} \\
\bottomrule

\end{tabular}
\end{adjustbox}
\label{tab:multimodal_models}
\end{table}

\subsubsection{Ablation on Different Image Encoders} We evaluate various image encoders to assess the quality of the extracted representations in multimodal training. We test \textit{ResNet 18} and \textit{ResNet 50} as CNN models, and 	\textit{ViT Tiny} and \textit{ViT Small} as transformer alternatives. We also use \textit{MedViT}, a hybrid model that captures local and global features. Notably, \textit{ViT Small} outperforms all others by at least 6\%. 

\subsubsection{Ablation on Key Parameters} Finally, we evaluate the impact of loss factor and random weight sampling to address class imbalance. Implementing a weighted random sampler significantly increases the model's performance by 39.6\% (from 36.1\% to 75.7\%). Furthermore, combining the sampling strategy with a loss factor of 1.2 applied to the positive class improves the F1-score by an additional 4.1\%, resulting in a final metric of 79.8\%. These results demonstrate the effectiveness of these strategies in managing imbalanced datasets.

\section{Limitations} 
Although the CARDIUM dataset represents a significant advance in automatic prenatal CHD diagnosis, several limitations remain. The limited number of CHD-positive cases in the first trimester and the overall small size of the dataset restrict the model’s ability to detect early-stage CHDs and to generalize effectively. Expanding the dataset is crucial for improving diagnostic performance. Furthermore, while generalization results are promising, the dataset’s exclusive focus on data from Colombian women may introduce demographic and geographic biases, underscoring the need for broader testing across diverse populations. Finally, variability in image quality and differences in how clinical protocols are applied by different specialists may impact real-world deployment, highlighting the need for multi-center validation.

\section{Conclusion}
In this work, we introduce CARDIUM, the first publicly available multimodal dataset for prenatal CHD detection, which integrates echocardiographic and ultrasound images with maternal clinical data. This dataset addresses the limitations associated with private datasets and unimodal approaches, providing a solid foundation for automated CHD diagnosis. Additionally, we propose a multimodal transformer architecture that leverages self-attention to capture intra-modality dependencies and cross-attention to model interactions between imaging and tabular features. Our model achieves an F1-score of 79.8\%, surpassing the image-only variation by 11\% and the tabular-only variation by 50\%, underscoring the advantages of multimodal integration for CHD detection. Moreover, our model generalizes well to an external ultrasound dataset, maintaining strong performance in unimodal multiclass classification. It also outperforms other multimodal state-of-the-art methods, which struggled to accurately detect CHD—likely due to the imbalanced nature of the dataset. These results demonstrate the robustness of our approach in imbalanced clinical scenarios.

\section{Acknowledgments}
Daniela Vega acknowledges the support of a UniAndes Google-DeepMind 2024 scholarship. 

\newpage

{
\small
\bibliographystyle{ieeenat_fullname}
\bibliography{main}
}


\clearpage
\appendix
\setcounter{page}{1}
\maketitlesupplementary

\setcounter{figure}{0}  
\renewcommand\thefigure{\Alph{figure}}

\begin{strip}
\centering
\includegraphics[width=\textwidth]{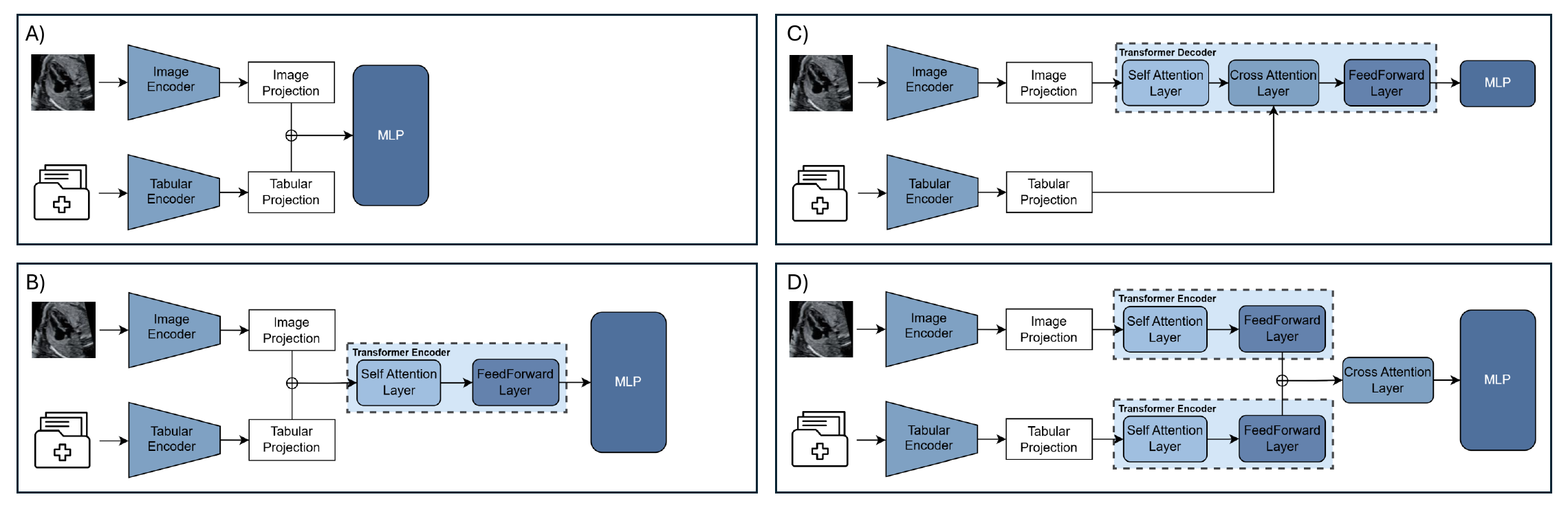}
\captionof{figure}{Comparison of multimodal fusion strategies. 
(A) \textit{MLP-Fusion}: concatenate modality features, then process them with an MLP.
(B) \textit{Transformer Encoder Fusion}: concatenate features, then process them with a transformer encoder.
(C) \textit{Transformer Decoder Fusion}: process image features with a decoder, then integrate tabular features through cross-attention.
(D) \textit{Transformer Encoder with Cross-Attention Fusion}: each modality is encoded separately, then fused via cross-attention.}
\label{fig:fusion_methods}
\end{strip}

\section{Implementation Details}
\subsection{Training and architecture of CARDIUM model} 
\label{sec:implemetation_details}

We train our model on an NVIDIA Quadro RTX 8000 and optimize parameters of the tabular, image, and multimodal module using Weights \& Biases \cite{wandb}. To address class imbalance, we employ loss weighting, image data augmentation, weighted random sampling, and hard positive mining (i.e., oversampling false negative examples). This last strategy was applied exclusively to the tabular encoder, where we apply a weighted random sampler on the trained loader every 20 epochs to oversample false negative examples. We train tabular and image encoders separately, freeze them, and then transfer the weights to the fusion module. We train our multimodal model for 100 epochs with binary cross-entropy loss, AdamW optimizer, and learning rate of $5 \times 10^{-7}$. The optimal multimodal parameters consist of eight-layer decoders with two attention heads and dropout rates of 0.4.

\subsection{Training on the External Ultrasound Fetal Dataset}
\label{sec:fetal-dataset}

To adapt our model for the external fetal ultrasound dataset, which is designed for image-only multiclass classification, we modify the classification head to output predictions for six classes and replace the binary cross-entropy loss with cross-entropy loss. Additionally, we optimize key hyperparameters to better suit the dataset’s larger size and more balanced class distribution. Specifically, we adjust the learning rate from  $5 \times 10^{-7}$ to  $4 \times 10^{-5}$ and reduce the dropout rates from 0.4 to 0.1. To evaluate the performance of our model pretrained on the CARDIUM dataset, we load the model's pretrained weights and modify the classification head, initializing it from scratch. We then finetune the model on the fetal dataset. Since we perform three-fold cross-validation, we finetune the best model for each fold, and during inference, we average the predictions from the three models to obtain the final prediction.

\subsection{Training TIP and MMCL on the CARDIUM Dataset}
\label{sec:sota}

We evaluate the performance of TIP and MMCL on the CARDIUM dataset, using the same fold and split distribution as the CARDIUM model to ensure a fair comparison. TIP was fine-tuned using publicly available pre-trained weights, originally trained on the UK Biobank \cite{ukbiobank2018}, which includes cardiac MRI images and clinical data. We followed the authors' recommended hyperparameters during fine-tuning. Since MMCL does not provide pre-trained weights, we trained it from scratch using the authors' suggested hyperparameters.

\subsection{Training with Half the Data}
\label{sec:ablation_half_data}

To train on half of the CARDIUM dataset, we split the training set in half while maintaining the same three-fold cross-validation setup, ensuring that each fold has a reduced training split. Additionally, we preserve the class and trimester distribution in the reduced training set to maintain consistency in data composition and allow for a fair comparison. The test split in each fold remained the same as in the original dataset, ensuring consistency in evaluation across all folds.

\section{Mathematical Formulation of Weight of Evidence Encoding} 
\label{sec:woeencoding}

For encoding categorical variables, we use Weight of Evidence (WoE) encoding combined with a five-fold cross-validation strategy. This technique can be summarized as follows,

\indent 
\begin{equation}
\text{WoE}_k(X) = \log \left( \frac{P(X \mid Y = 1, D_{-k})}{P(X \mid Y = 0, D_{-k})} \right)
\end{equation}
\indent

where $\text{WoE}_k(X)$ denotes the Weight of Evidence value for category $X$ in fold $k$; $P(X \mid Y = 1,\ D_{-k})$ is the probability of observing $X$ among positive samples in the data excluding fold $k$; $P(X \mid Y = 0,\ D_{-k})$ is the probability of observing $X$ among negative samples in the data excluding fold $k$; and $D_{-k}$ represents the dataset excluding fold $k$.

\section{Architecture of the Different Multimodal Fusion Strategies}
\label{sec:multimodal_fusion}

The different multimodal fusion strategies implemented are depicted in Figure \ref{fig:fusion_methods}. The MLP Fusion strategy takes the output of each modality encoder, concatenates the features, and then processes them with an MLP. The Transformer Encoder Fusion strategy concatenates the modality features and processes them with a transformer encoder. The resulting output is then passed through an MLP. The Transformer Decoder Fusion strategy processes the image features with a transformer decoder and integrates the tabular features through the cross-attention layer. The output is then processed by an MLP. Finally, the Transformer Encoder with Cross-Attention Fusion strategy processes the features of each modality separately with its own transformer encoder. The outputs of these encoders are fused using a cross-attention layer and then processed with an MLP.

\end{document}